\documentclass[11pt,a4paper]{article}
\usepackage[hyperref]{naaclhlt2019}
\usepackage{times}
\usepackage{latexsym}

\usepackage{url}

\usepackage{graphicx}
\usepackage{amsmath}
\usepackage{multirow}
\usepackage{bbold}
\usepackage{algorithm} 
\usepackage{algorithmicx}
\usepackage{algpseudocode}
\usepackage{xcolor}

\usepackage{array}
\newcolumntype{L}[1]{>{\raggedright\let\newline\\\arraybackslash\hspace{0pt}}m{#1}}
\newcolumntype{C}[1]{>{\centering\let\newline\\\arraybackslash\hspace{0pt}}m{#1}}
\newcolumntype{R}[1]{>{\raggedleft\let\newline\\\arraybackslash\hspace{0pt}}m{#1}}

\aclfinalcopy 
\def\aclpaperid{2219} 

\title{Fact Discovery from Knowledge Base via Facet Decomposition}

  \author{Zihao Fu$^{1}$, Yankai Lin$^{2}$, Zhiyuan Liu$^{2}$\thanks{\ Corresponding author: Zhiyuan Liu (liuzy@tsinghua.edu.cn).}, Wai Lam$^{1}$ \\
$^{1}$  Department of Systems Engineering and Engineering Management \\ The Chinese University of Hong Kong, Hong Kong \\
$^{2}$ Department of Computer Science and Technology, \\State Key Lab on Intelligent Technology and Systems, \\
National Lab for Information Science and Technology, Tsinghua University, Beijing, China \\
}

\date{}

\begin{document}
\maketitle
\begin{abstract}
During the past few decades, knowledge bases (KBs) have experienced rapid growth. Nevertheless, most KBs still suffer from serious incompletion. Researchers proposed many tasks such as knowledge base completion and relation prediction to help build the representation of KBs. However, there are some issues unsettled towards enriching the KBs. Knowledge base completion and relation prediction assume that we know two elements of the fact triples and we are going to predict the missing one. This assumption is too restricted in practice and prevents it from discovering new facts directly. To address this issue, we propose a new task, namely, fact discovery from knowledge base. This task only requires that we know the head entity and the goal is to discover facts associated with the head entity. To tackle this new problem, we propose a novel framework that decomposes the discovery problem into several facet discovery components. We also propose a novel auto-encoder based facet component to estimate some facets of the fact. Besides, we propose a feedback learning component to share the information between each facet. We evaluate our framework using a benchmark dataset and the experimental results show that our framework achieves promising results. We also conduct extensive analysis of our framework in discovering different kinds of facts. The source code of this paper can be obtained from \url{https://github.com/thunlp/FFD}.

\end{abstract}

\section{Introduction}

\begin{figure}[!t]
\centering
\includegraphics[width=1.0\columnwidth]{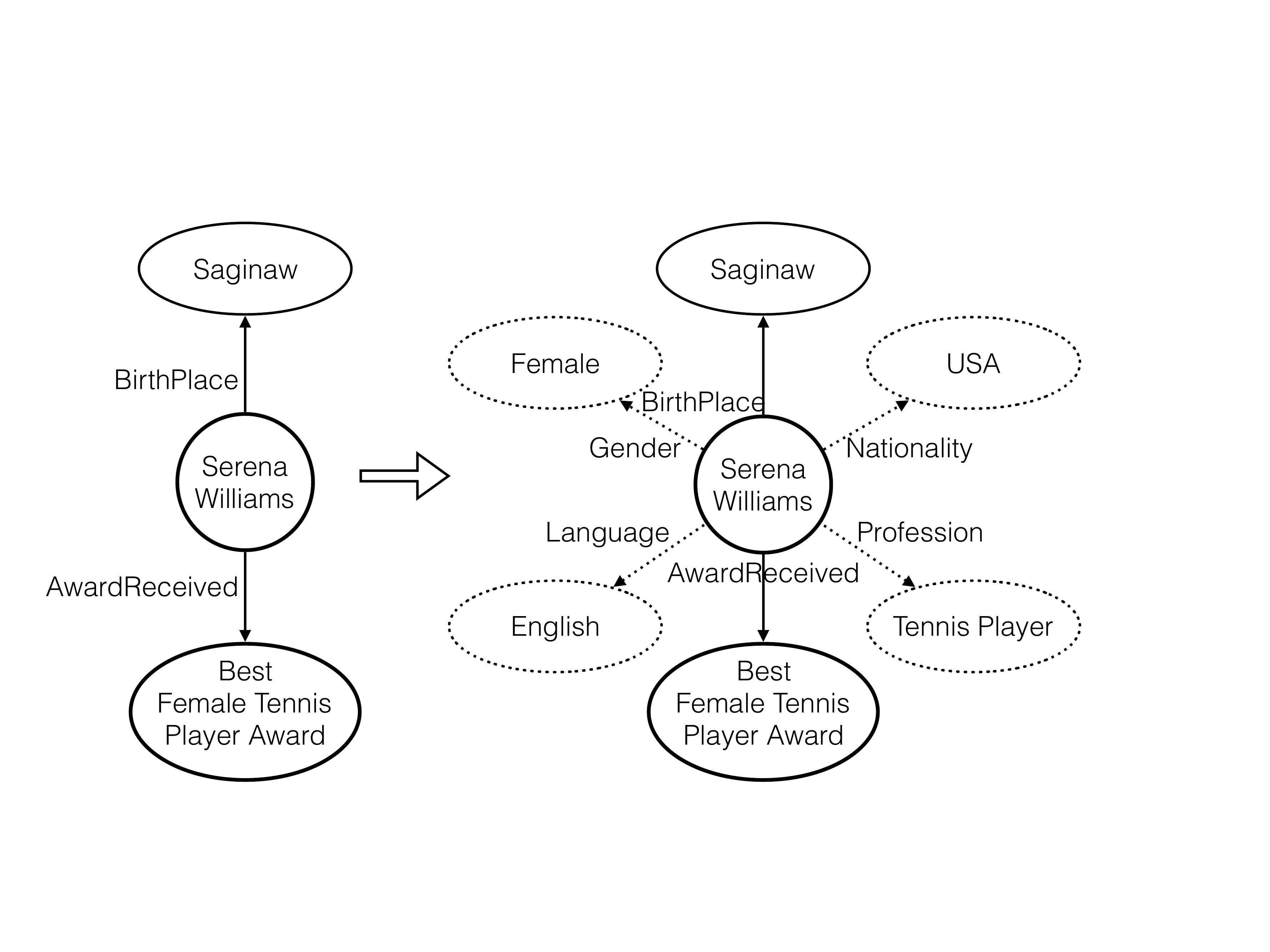}
\caption{In the FDKB task, only the head entity is given. The relation and tail entity should be discovered simultaneously given the head entity.}
\label{fig:task}
\end{figure}

Recent years have witnessed the emergence and growth of many large-scale knowledge bases (KBs) such as Freebase \cite{bollacker2008freebase}, DBpedia \cite{lehmann2015dbpedia}, YAGO \cite{suchanek2007yago} and Wikidata \cite{vrandevcic2014wikidata} to store facts of the real world. Most KBs typically organize the complex structured information about facts in the form of triples (\emph{head entity}, \texttt{relation}, \emph{tail entity}), e.g., (\emph{Bill Gates}, \texttt{CEOof}, \emph{Microsoft Inc.}). These KBs have been widely used in many AI and NLP tasks such as text analysis \cite{berant2013semantic}, question answering \cite{bordes2014question}, and information retrieval \cite{hoffmann2011knowledge}.

The construction of these KBs is always an ongoing process due to the endless growth of real-world facts. Hence, many tasks such as knowledge base completion (KBC) and relation prediction (RP) are proposed to enrich KBs.

The KBC task usually assumes that one entity and the relation $r$ are given, and another entity is missing and required to be predicted. In general, we wish to predict the missing entity in $(h, r, ?)$ or $(?, r, t)$, where $h$ and $t$ denote a head and tail entity respectively. Similarly, the RP task predicts the missing relation given the head and tail entities and their evidence sentences, i.e. filling $(h, ?, t)$. Nevertheless, the assumption of knowing two parts of the triple is too strong and is usually restricted in practice.

In many cases, we only know the entity of interest, and are required to predict both its attributive relations and the corresponding entities. As shown in Figure \ref{fig:task}, the task is to predict the fact triples when given only the head entity, i.e. filling $(h,?,?)$. Since any entity can serve as the head entity for identifying its possible fact triples, this task should be more practical for real-world settings. This task is non-trivial since less information is provided for prediction. We name the task as Fact Discovery from Knowledge Base (FDKB).

Some existing methods such as knowledge base representation (KBR) can be applied to tackle the FDKB task with simple modifications. KBR models typically embed the semantics of both entities and relations into low-dimensional semantic space, i.e., embeddings. For example, TransE \cite{bordes2013translating} learns low-dimensional and real-valued embeddings for both entities and relations by regarding the relation of each triple fact as a translation from its head entity to the tail entity. TransE can thus compute the valid score for each triple by measuring how well the relation can play a translation between the head and tail entities. Many methods have been proposed to extend TransE to deal with various characteristics of KBs \cite{jiknowledge2015,ji2016knowledge,he2015learning,lin2015modeling}.

To solve the FDKB task using KBR, one feasible way is to exhaustively calculate the scores of all $(r, t)$ combinations for the given head entity $h$. Afterwards, the highly-scored facts are returned as results. However, this idea has some drawbacks: (1) It takes all relations to calculate ranking scores for each head entity, ignoring the nature of the head entity. The combination of all possible relations and tail entities will lead to huge amount of computations. (2) A large set of candidate triples immerses the correct triples into a lot of noisy triples. Although the probability of invalid facts getting a high score is small, with the large size of the candidate set, the total number of invalid facts with high score is non-negligible.

To address the above issues, we propose a new framework named as fact facet decomposition (FFD). The framework follows human being's common practice to identify unknown facts: One typically firstly investigates which relation that a head may have, and then predicts the tail entity based on the predicted relation. This procedure actually utilizes information from several perspectives. Similarly, FFD decomposes fact discovery into several facets, i.e., head-relation facet, tail-relation facet, and tail inference facet, and model each facet respectively. The candidate fact is considered to be correct when all of the facets are trustworthy. We propose a novel auto-encoder based entity-relation component to discover the relatedness between entities and relations. Besides, we also propose a feedback learning component to share the information between each facet.

We have conducted extensive experiments using a benchmark dataset to show that our framework  achieves promising results. We also conduct an extensive analysis of the framework in discovering different kinds of facts. The contributions of this paper can be summarized as follows: (1) We introduce a new task of fact discovery from knowledge base, which is more practical. (2) We propose a new framework based on the facet decomposition which achieves promising results. 

\section{Related Work}

In recent years, many tasks \cite{wang2017knowledge} have been proposed to help represent and enrich KBs. Tasks such as knowledge base completion (KBC) \cite{bordes2013translating,wang2014knowledge,jiknowledge2015,ji2016knowledge,wang2017knowledge} and relation prediction (RP) \cite{mintz2009distant,lin2015modeling,xie2016representation} are widely studied and many models are proposed to improve the performance on these tasks. However, the intention of these tasks is to test the performance of models in representing KBs and thus they cannot be used directly to discover new facts of KBs.
Moreover, our FDKB task is not a simple combination of the KBC and RP task since both of these two tasks require to know two of the triples while we assume we only know the head entity.

A common approach to solving these tasks is to build a knowledge base representation (KBR) model with different kinds of representations. Typically, one element of the triples is unknown. Then, all entities are iterated on the unknown element and the scores of all combinations of the triples are calculated and then sorted. 
Many works focusing on KBR  attempt to encode both entities and relations into a low-dimensional semantic space. KBR models can be divided into two major categories, namely translation-based models and semantic matching models \cite{wang2017knowledge}.

Translation-based models such as TransE \cite{bordes2013translating} achieves promising performance in KBC with good computational efficiency. TransE regards the relation in a triple as a translation between the embedding of head and tail entities. It means that TransE enforces that the head entity vector plus the relation vector approximates the tail entity vector to obtain entity and relation embeddings. However, TransE suffers from problems when dealing with 1-to-N, N-to-1 and N-to-N relations. To address this issue, TransH \cite{wang2014knowledge} enables an entity to have distinct embeddings when involving in different relations. TransR \cite{lin2015learning} models entities in entity space and uses transform matrices to map entities into different relation spaces when involving different relations. Then it performs translations in relation spaces. In addition, many other KBR models have also been proposed to deal with various characteristics of KBs, such as TransD \cite{jiknowledge2015}, KG2E \cite{he2015learning}, PTransE \cite{lin2015modeling}, TranSparse \cite{ji2016knowledge}.

Semantic matching models such as RESCAL \cite{nickel2011three}, DistMult\cite{yang2014embedding}, Complex \cite{trouillon2016complex}, HolE \cite{nickel2016holographic} and ANALOGY \cite{liu2017analogical} model the score of triples by the semantic similarity. RESCAL simply models the score as a bilinear projection of head and tail entities. The bilinear projection is defined with a matrix for each relation. However, the huge amount of parameters makes the model prone to overfitting. To alleviate the issue of huge parameter space, DistMult is proposed to restrict the relation matrix to be diagonal. However, DistMult cannot handle the asymmetric relations. To tackle this problem, Complex is proposed assuming that the embeddings of entities and relations lie in the space of complex numbers. This model can handle the asymmetric relations. Later, Analogy is proposed by imposing restrictions on the matrix rather than building the matrix with vector. It achieves the state-of-the-art performance. Besides,  \cite{bordes2011learning,socher2013reasoning,chen2013learning,bordes2014semantic,dong2014knowledge,liu2016probabilistic} conduct the semantic matching with neural networks. An energy function is used to jointly embed relations and entities.

\section{Problem Formulation}

We denote $\mathcal{E}$ as the set of all entities in KBs, $\mathcal{R}$ is the set containing all relations. $|\mathcal{E}|$ and $|\mathcal{R}|$ stand for the size of each set respectively. A fact is a triple $(h,r,t)$ in which $h,t\in \mathcal{E}$ and $r\in \mathcal{R}$. $\mathcal{T}$ is the set of all true facts.

When a head entity set $\mathcal{H}$ is given, a new fact set is to be discovered based on these head entities. The discovered fact set is denoted as $\mathcal{T}_{d}=\{(h,r,t)|h\in\mathcal{H}\}$. Our goal is to find a fact set $\mathcal{T}_{d}$ that maximizes the number of correct discovered facts:
\begin{equation}
\begin{aligned}
\max_{\mathcal{T}_{d}} &|\mathcal{T}_{d}\cap \mathcal{T}|\\
\text{s.t. }& |\mathcal{T}_{d}| = K,
\end{aligned}
\label{equ:primal}
\end{equation}
in which $K$ is a user-specified size.

\section{Methodology}
\subsection{Fact Facet Decomposition Framework}
Problem (\ref{equ:primal}) is intractable since the set $\mathcal{T}$ is unknown. We tackle this problem by estimating a fact confidence score function $c(h,r,t)$ for each fact in $\mathcal{T}_{d}$ and maximizing the total score. The problem is then formulated as:

\begin{equation}
\begin{aligned}
\max_{\mathcal{T}_{d}} ~&\sum_{(h,r,t)\in \mathcal{T}_{d}}c(h,r,t)\\
\text{s.t. }& |\mathcal{T}_{d}| = K.
\end{aligned}
\label{equ:newproblem}
\end{equation}

To integrate the information from various facets of the fact, our framework, known as Fact Facet Decomposition (FFD) framework, decomposes the fact discovery problem into several facet-oriented detection tasks. A fact is likely to be correct if all facets provide supportive evidence. The facets are as follows:

\begin{enumerate}
\item Head-relation facet: A fact is likely true, if the head entity has a high probability of containing the relation. This is denoted as $f_h(r)$; 

\item Tail-relation facet: A fact is likely true, if the tail entity has a high probability of containing the relation. This is denoted as $f_t(r)$; 
\item Tail inference facet: A fact is likely true, if the score of the tail entity is high with respect to the given head and relation. This is denoted as $f_{h,r}(t)$.
\end{enumerate}
Therefore, the facet confidence score can be expressed as: 
\begin{equation}
c(h,r,t)=\lambda_1 f_h(r)+\lambda_2 f_t(r)+\lambda_3 f_{h,r}(t),
\end{equation}
where $\lambda_1,\lambda_2,\lambda_3$ are weight parameters. The head-relation facet and the tail-relation facet can be both modeled with an entity-relation facet component. The tail inference facet can be modeled by a KBR component.

\begin{figure}[!t]
\centering
\includegraphics[width=1.0\columnwidth]{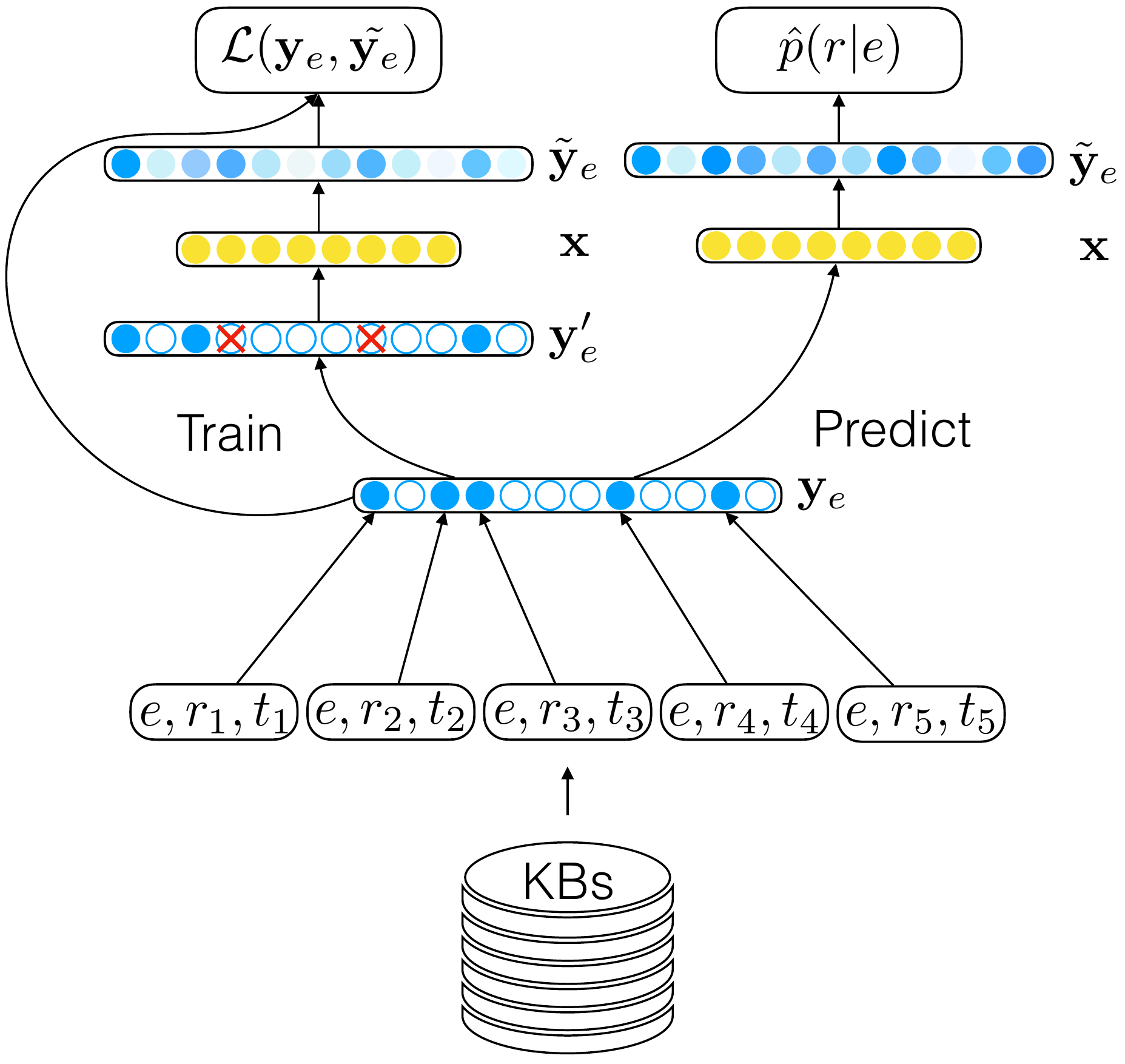}
\caption{The structure of the entity-relation component.}
\label{fig:framework}
\end{figure}

\subsubsection{Entity-relation Facet Component}
The entity-relation component estimates the probability of a relation given an entity. The structure is shown in Figure \ref{fig:framework}. It is modeled as the logarithm of the estimated conditional probability: 
\begin{equation}
f_e(r)=\log \hat{p}(r|e),
\end{equation}
where $e=h$ or $t$. $\hat{p}(r|e)$ aims at measuring the probability of a relation that this entity may have. In order to estimate this probability, the existing relations of a head or tail entity is used to infer other related relations. For example, if a head entity has an existing fact in which the relation is ``BirthPlace'', we may infer that this head entity may be a person and some relations such as ``Gender'', ``Language'' may have a high probability of association with this head entity. Therefore, the problem is transformed into a problem that estimates the relatedness between relations. To infer the probability of each relation based on existing relations, we employ a denoising auto-encoder \cite{vincent2008extracting} which can recover almost the same representation for partially destroyed inputs. Firstly, facts related to an entity is extracted from the KBs. Then, this entity is encoded by the existing relations. Let $\mathbf{y}_e\in \mathbb{R}^{|\mathcal{R}|}$ be the 0-1 representation of relations that $e$ has. $\mathbf{y}_{ei}$ indicates whether the entity $e$ has the relation $i$ or not. During the training phase, non-zero elements in $\mathbf{y}_e$ is randomly set to zero and the auto-encoder is trained to recover the corrupted elements. The corrupted vector is denoted as $\mathbf{y}_e'$.

Formally, our structure encoder first maps the corrupted one-hot vector $\mathbf{y}_e'$ to a hidden representation $\mathbf{x}\in \mathbb{R}^{d_1}$ of the entity through a fully connected layer:
\begin{equation}
  \mathbf{x} = \tanh(\mathbf{W}_f\mathbf{y}_e'+\mathbf{b}_f),
\end{equation}
where $\mathbf{W}_f\in \mathbb{R}^{d_1\times |\mathcal{R}|}$ is the translation matrix and $\mathbf{b}_f\in \mathbb{R}^{d_1}$ is the bias vector. $\mathbf{x}$ is the vector representation of  the entities in a hidden semantic space. In this space, similar entities are close to each other while entities of different types are far from each other. If some relations are missing, the fully connected layer will also map the entity into a nearby position. 

Afterwards, $\mathbf{x}$ is used to recover the probability distribution for all relations through a fully connected layer and a sigmoid layer:
\begin{equation}
  \tilde{\mathbf{y}}_e = \text{sigmoid}(\mathbf{W}_g\mathbf{x}+\mathbf{b}_g),
\end{equation}
where  $\mathbf{W}_g\in \mathbb{R}^{|\mathcal{R}|\times d_1}$ and $\mathbf{b}_g\in \mathbb{R}^{|\mathcal{R}|}$ is the weight matrix and bias vector of the reverse mapping respectively. $\tilde{\mathbf{y}}_e$ is the recovered probability distribution of each relation (therefore, the sum of each element in $\tilde{\mathbf{y}}_e$ does not necessarily equal to 1). This layer will map the entity representation in the semantic space into a probability vector over all relations. Since similar entities are located in the adjacent area, they are likely to have a similar relation probability. Therefore, the probability of missing relations will also be high though the relations are unknown.

We use the original one-hot representation of the relations and the recovered relation probability to calculate a loss function:

\begin{eqnarray}
\mathcal{L}(\mathbf{y}_e,\tilde{\mathbf{y}}_e)=-\sum_{e=1}^{|\mathcal{E}|} \sum_{i=1}^{|\mathcal{R}|}\{ \mathbf{y}_{ei}\log(\tilde{\mathbf{y}}_{ei})+\nonumber \\
(1-\mathbf{y}_{ei})\log(1-\tilde{\mathbf{y}}_{ei})\}.
\end{eqnarray}
The loss function forces the output $\tilde{\mathbf{y}}_{ei}$ to be consistent with $\mathbf{y}_{ei}$ which makes it capable of discovering all related relations from known relations. It can be optimized with an Adam \cite{kingma2014adam} based optimizer.

When predicting new facts, the one-hot representation $\mathbf{y}_e$ is sent into the auto-encoder directly instead of using the corrupted representation. The result $\tilde{\mathbf{y}}_e$ is the estimated probability of each relation, i.e. 
\begin{equation}
\hat{p}(r=i|e)=\tilde{\mathbf{y}}_{ei}.
\end{equation}
This probability will be high if the relation $i$ is closely related to the existing relations of the entity $e$.

\subsubsection{Tail Inference Facet Component}
We use a KBR component to model the tail inference facet $f_{h,r}(t)$. Three KBR models are investigated namely DistMult, Complex, and Analogy.

The DistMult model defines the score function as $f_r(h,t)=\mathbf{h}^T\text{diag}(\mathbf{r})\mathbf{t}$, in which $\mathbf{h}, \mathbf{r}, \mathbf{t}$ are vector representation of the head, relation and tail respectively. The learning objective is to maximize the margin between true facts and false facts.

It can decrease the score of the wrong facts and increase the score of the true facts at the same time.

The Complex model employs complex number as the KBR embedding. Therefore, the score function is defined as $f_r(h,t)=\text{Re}(\mathbf{h}^T\text{diag}(\mathbf{r})\bar{\mathbf{t}})$, in which $\mathbf{h}, \mathbf{r}, \mathbf{t}$ are complex vectors and $\bar{\mathbf{t}}$ stands for the conjugate of $\mathbf{t}$.

The Analogy model does not restrict the relation matrix to be diagonal. Therefore, the score function is $f_r(h,t)=\mathbf{h}^T\mathbf{M_r}\mathbf{t}$, in which $\mathbf{M_r}$ is the matrix corresponding to the relation $r$. Since many relations satisfy normality and commutativity requirements, the constraints can thus be set as $W_rW_r^T=W_r^TW_r,\forall r\in \mathcal{R}$ and $W_rW_{r'}=W_{r'}W_r,\forall r,r' \in \mathcal{R}$. Solving such a problem is equivalent to optimizing the same objective function with the matrix constrained to almost-diagonal matrices\cite{liu2017analogical}.

After the score function is calculated, the tail inference facet $f_{h,r}(t)$ is modeled by a softmax function:
\begin{equation}
f_{h,r}(t)=\log \hat{p}(t|h,r)=\log \frac{e^{f_r(h,t)}}{\sum_{t'\in\mathcal{E}} e^{f_{r}(h,t')}}.
\end{equation}
It should be noted that the normalization step is only conducted on the tail entities since the head and relation are the input of the model. We only use these three models due to the limited space. Other models can be embedded into our framework easily in the same way.

\subsection{Fact Discovery Algorithm}

As mentioned above, we need to calculate $f_h(r)$, $f_t(r)$ and $f_{h,r}(t)$. $f_h(r)$ and $f_t(r)$ are computed by the entity-relation component while $f_{h,r}(t)$ is computed by the tail inference component. Recall that a fact is likely to be true when all the facets exhibit strong support. In other words, we can prune away the fact if one of the facets is low and stop calculating other facets. Based on this strategy, we design two additional constraints on Problem (\ref{equ:newproblem}). Therefore, this method can be viewed as a shrink of the constraint space of the optimization problem. The new problem can be expressed as:
\begin{equation}
\begin{aligned}
\max_{\mathcal{T}_{d}} ~&\sum_{(h,r,t)\in \mathcal{T}_{d}}\{\lambda_1 f_h(r)+\lambda_2 f_t(r)+\lambda_3 f_{h,r}(t)\}\\
\text{s.t. } &h\in\mathcal{H};|\mathcal{T}_{d}|=K\\
&f_h(r)>\tau_h;~\sum_{r}\mathbb{1}(f_h(r) > \tau_h)=n_h\\
&f_t(r)>\tau_t;~\sum_{r}\mathbb{1}(f_t(r) > \tau_t)=n_t,
\end{aligned}
\label{equ:filter}
\end{equation}
where $\mathbb{1}_A(x)$ is an indicator function. $\mathbb{1}_A(x)=1$ if $x\in A$ and $\mathbb{1}_A(x)=0$ otherwise. $n_h$ and $n_t$ are the user-specified parameters indicating top-$n_h$ or top-$n_t$ relations are considered. $\lambda_1,\lambda_2$ and $\lambda_3$ are fixed hyperparameters.

Problem (\ref{equ:filter}) is actually a mixed integer linear programming problem. We start to solve this problem from the constraints. Since $f_t(r)$ is independent of the given $\mathcal{H}$, it can be preprocessed and can be reused for other queries. When a head entity $h$ is given, we firstly calculate $f_h(r)$ and get top-$n_h$ relations ranked by $f_h(r)$. Then, for each relation, $f_t(r)$ is used to get the top-$n_t$ entities. Afterwards, the tail inference facet $f_{h,r}(t)$ will be calculated for all remaining relations and entities and top-$n_f$ triples will be cached. Finally, top-$\bar{K}$ facts ranked by the facet confidence score $c(h,r,t)$ is returned as the new facts discovered for the entity $h$, where $\bar{K}=K/|\mathcal{H}|$ stands for the average fact number for each head entity.

\subsection{Feedback Learning} The three facets depict the characteristics of the KBs from different perspectives. For example, the head-relation facet indicates which relation the head entity may have. The tail-relation facet can be interpreted in a similar manner. We propose a feedback learning (FL) component for the facets to share the information in different perspectives with each other. FL feeds the predicted facts back to the training set to enhance the training procedure and iterates the process of predicting and training several times. In each iteration, the information from different perspectives is shared with each facet via the newly added facts.

Specifically, after predicting the top-$n_h$ facts for each head entity, we select top-$n_{fb}$ ($n_{fb}<n_h$) most probable facts according to the score of each triple and then feed them into the existing knowledge base for re-training the FFD model. We repeat the above updating operation several rounds.

\section {Experiment}
\subsection{Dataset}
We evaluate our framework by re-splitting a widely used dataset FB15k \cite{bordes2013translating}, which is sampled from Freebase. It contains $1,345$ relations and $14,951$ entities.

In FB15k, some of the testing set's head entities do not appear in the training set as head entities. To evaluate our framework, we construct the new dataset. We re-split FB15k into training ($\mathcal{T}_{train}$), validation ($\mathcal{T}_{valid}$) and testing ($\mathcal{T}_{test}$) set, and make sure that there is no overlap between the three sets. For all head entities in $\mathcal{H}$, a relation ratio $R\%$ is used to assign the facts into training and testing set. $R\%$ relations of a head entity are in the training set while the other $1-R\%$ are in the testing set. In order to evaluate the task, we require that the head entities in $\mathcal{H}$ is the same as the testing head entity and is a subset of the training head set, i.e. $\mathcal{H} = \{h|(h,r,t)\in \mathcal{T}_{test},\exists r,t \in \mathcal{E}\}  \subset \{h|(h,r,t)\in \mathcal{T}_{train},\exists r,t\in \mathcal{E}\}$. We set $R=50$. After the splitting, the training, testing and validation set size is $509,339$, $41,861$ and $41,013$ respectively.

\subsection{Comparison Models}
To demonstrate the effectiveness of our framework, we provide several strong comparison models that can be used for solving this task.

\subsubsection{Matrix Factorization Models (SVD and NMF)} MF models firstly count the frequency of all relation-tail pairs. Some low-frequency relation-tail pairs are ignored to save computational time. Afterwards, we build a (head, relation-tail) co-occurrence matrix $M^C \in \mathbb{R}^{|\mathcal{E}|\times p}$, in which $p$ is the size of the relation-tail pair set. Each element $M^C_{ij}$ in the matrix represents whether the head entity $i$ has the relation-tail pair $j$ or not. Then, the matrix will be decomposed by the product of two matrices, i.e.
\begin{equation}
M^C\approx WH,
\end{equation}
in which $W\in \mathbb{R}^{|\mathcal{E}|\times k}, H\in \mathbb{R}^{k\times p}$. $k$ is the hidden category number of the head and relation-tail pairs. The decomposition can be achieved in several ways with different assumptions. Two kinds of matrix decomposition models are used namely SVD \cite{halko2011finding} and NMF \cite{lee1999learning}.

In the prediction stage, a new matrix is constructed by $M'^C=WH$. For each row in $M'^C$, we record top-$\bar{K}$ relation-tail pairs and their scores. The MF models always suffer from the sparsity problem since a lot of relation-tail pairs are ignored.

\subsubsection{KBR+ Models (DistMult+, Complex+ and Analogy+)}
The most straightforward method of estimating the fact confidence score $c(h,r,t)$ is to use KBR model directly to evaluate each triples' score. We exhaustively score all possible combinations of relations and tails and use the highly-scored facts to make up the set $\mathcal{T}_{d}$. We select some state-of-the-art models including DistMult \cite{yang2014embedding}, Complex \cite{trouillon2016complex} and Analogy \cite{liu2017analogical}. We denote them as DistMult+, Complex+ and Analogy+.

After a KBR model learns a score function $f_r(h,t)$, the probability of each $(r,t)$ pair with respect to a given head entity can be estimated by a softmax function:
\begin{equation}
\hat{p}(r,t|h)=\frac{e^{f_r(h,t)}}{\sum_{r'\in \mathcal{R}} \sum_{t'\in\mathcal{E}} e^{f_{r'}(h,t')}}.
\end{equation}
Afterwards, the score of each fact is sorted and top-$\bar{K}$ relation-tail pairs for a head entity are regarded as the predicted results.

\subsection{Experimental Setup}
There are 2,000 head entities in the testing set. Therefore, we predict the corresponding relation and tail entity with respect to these 2,000 head entities. In MF models, only relation-tail pairs that occur more than 3 times in the training set are considered (24,615 pairs in total). For each head entity, we set $\bar{K}=50$. In KBR+, we also set $\bar{K}=50$. For our framework, we set $n_h=n_t=30$, $n_f=10$, $\bar{K}=50$, $\lambda_1=1.0,\lambda_2=1.0,\lambda_3=0.5$. The auto-encoder iterates for 1,000 epochs and the learning rate for Adam is 0.005. For the feedback learning component, we set $n_{fb}=20,000$. With this setting, each model returns 100,000 facts.

We use four evaluation metrics, including precision, recall MAP, and F1 in relation prediction. Precision is defined as the ratio of the true positive candidates' count over the number of all the retrieved candidates' count. Recall is defined as the ratio of the true positive candidates' count over all the positive facts' count in the testing set. MAP \cite{manning2008introduction} is a common evaluation method in information retrieval tasks. F1 is defined as the harmonic mean of the precision and recall.

\subsection{Experimental Results}
The experimental result is shown in Table \ref{tab:main}. From the experiment result, we observe the followings:

\begin{table}[!h]

\centering
\small
\begin{tabular}{C{2.0cm}cccc}
\hline
Method & MAP & precision & recall & F1 \\
\hline
SVD & 0.0873 & 0.0897 & 0.2143 & 0.1265  \\
NMF & 0.0827 & 0.0857 & 0.2048 & 0.1209  \\
DistMult+ & 0.1086 & 0.1068 & 0.2552 & 0.1506  \\
Complex+  & 0.2384 & 0.1608 & 0.3842 & 0.2267  \\
Analogy+  & 0.2367 & 0.1606 & 0.3837 & 0.2265  \\
\hline
FFD (DistMult) & 0.2486 & 0.1939 & 0.4633 & 0.2734  \\
FFD (Complex)  & 0.2723 & 0.1991 & 0.4758 & 0.2808  \\
FFD (Analogy)  & \textbf{0.2769} & \textbf{0.2001} & \textbf{0.4779} & \textbf{0.2821}  \\
\hline
FFD (Analogy) w/o FL & 0.2308 & 0.1978 & 0.4725 & 0.2788 \\
\hline
\end{tabular}
\caption{Results of our framework and comparison models.}
\label{tab:main}
\end{table}

\begin{enumerate}
\item FFD based model outperforms other models in all metrics. It illustrates the advantage of our decomposition design. Moreover, in FFD, using Analogy to predict $c(h,r,t)$ outperforms Complex. One reason is that the discovery algorithm harnesses the relatively large parameter space of Analogy and avoids some occasionally emerging wrong facts.
\item The relation of the head entity can be correctly predicted. This is because, in training, we remove some relations and the auto-encoder is trained to learn to recover the missing relations based on the remaining relations.

\item The MF based models (i.e. SVD and NMF) do not perform as good as KBR+ models and FFD. The reason is partially due to the sparsity problem in MF models. A lot of relation-tail pairs have not been used as the feature and thus cannot be predicted.

\item Different from the traditional KBC task, Complex performs slightly better than Analogy. One reason is that Analogy's constraint is looser than Complex. Therefore, it may easily predict wrong facts due to error propagation.

\item The ablation experiment shows that the feedback learning can improve the performance effectively.
\end{enumerate}

To illustrate the capability of handling different kinds of relations, we plot the accuracy with respect to different kinds of relations. We use heads per tail (hpt) and tails per head (tph) index to represent the difficulty of each relation. If the relation's index is high, it means that each head of the relation may have more tails or vice versa. These relations are more difficult to predict. This is the similar problem of 1-N, N-1 and N-N relation in KBC task. The plot is shown in Figure \ref{fig:result}. From the figure, we can observe the followings:

\begin{figure}[!t]
\centering
\includegraphics[width=1.0\columnwidth]{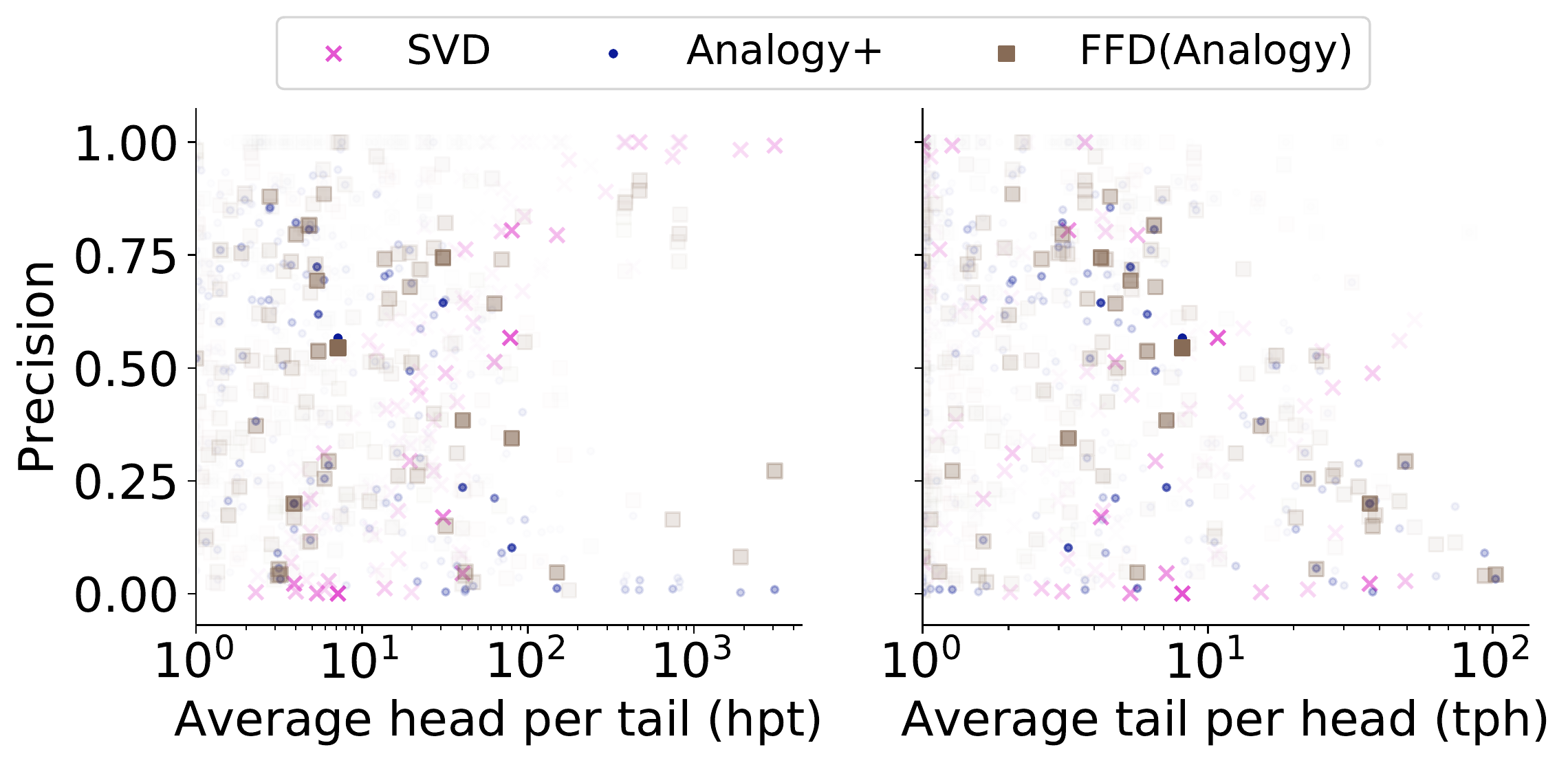}
\caption{Precision on each relation. Deep color stands for the fact that number of fact is large.}
\label{fig:result}
\end{figure}

\begin{enumerate}
\item FFD can be adapted to all kinds of relations with different hpt and tph.
\item MF, KBR+, FFD models can handle relations with relatively high hpt but fail with high tph. This is because our goal is to predict the relation and the tail based on the head. Therefore, the choice may be harder to make with high tph.
\item As the hpt grows, the precision of SVD model also grows. The reason is that as hpt grows, the sparsity problem is alleviated. Therefore, the performance of SVD grows.
\end{enumerate}

\subsection{Sparsity Investigation}
The MF model suffers from the sparsity problem since a lot of relation-tail pairs do not appear in the training set. We examine the training set and observe that 97.46\% relation-tail pairs do not appear and 0.34\% relation-tail pairs appear for only one time. These pairs can hardly provide any information for the MF models either.

\begin{table}[!h]
\centering
\small
\begin{tabular}{cccc}
\hline
Relation Ratio & train & test & valid \\
\hline
10\% & 451,214 & \multirow{5}{*}{41,861} & \multirow{5}{*}{41,013} \\
20\% & 462,395 &  &  \\
30\% & 475,841 &  &  \\
40\% & 491,498 &  &  \\
50\% & 509,339 &  &  \\
\hline
\end{tabular}
\caption{Statistics of dataset with different relation ratio.}
\label{tab:dataset}
\end{table}

To test whether our framework is capable of dealing with the data sparsity problem. We remove training facts which contains head entities in $\mathcal{H}$ according to a specific ratio. We decrease the relation ratio $R\%$ ranging from $50\%$ to $10\%$ to explore the effectiveness of our framework in discovering new facts. The dataset statistics is shown in Table \ref{tab:dataset}. We apply FFD (Analogy) on each dataset. As shown in Table \ref{tab:sparsity}, precision, F1 and recall decrease since the data becomes more and more sparse. MAP increases slightly since it is averaged on all extracted facts. When the number of extracted facts decreases, some facts ranked at the bottom with low scores are excluded.

\begin{table}[!h]
\centering
\small
\begin{tabular}{ccccccc}
\hline
Relation Ratio & MAP & precision & recall & F1 \\
\hline
50\% & 0.2101 & 0.1851 & 0.4421 & 0.2609 \\
40\% & 0.2099 & 0.1768 & 0.4224 & 0.2493 \\
30\% & 0.2167 & 0.1686 & 0.4029 & 0.2378 \\
20\% & 0.2236 & 0.1623 & 0.3878 & 0.2289 \\
10\% & 0.2497 & 0.1534 & 0.3664 & 0.2162 \\
\hline
\end{tabular}
\caption{Result of dataset in different relation ratio.}
\label{tab:sparsity}
\end{table}

\subsection{Case Study}
We provide a case study to demonstrate the characteristics of different models and show that our FFD can utilize more information. We choose the head entity ``Stanford Law School'' (Freebase ID: /m/021s9n). The predicted facts of SVD, Analogy+ and FFD (Analogy) are shown in Table \ref{tab:casestudy}. From the table, we can observe the followings:

\begin{table}[!h]
\centering
\small
\begin{tabular}{C{1.8cm}C{2.0cm}|C{0.3cm}|C{0.4cm}C{0.5cm}C{0.5cm}}
\hline
Relation 				&Tail 				&In RT pair 	&SVD 		&Analog+ 	&FFD (Analogy) \\
\hline
Located In				&USA				&${\surd}$		&${\surd}$	&			&${\surd}$ \\
Located In				&California			&${\surd}$		&			&${\surd}$	&${\surd}$ \\
Located In				&Stanford			&				&			&			&${\surd}$ \\
Educational Institution	&Stanford Law School	&				&			&${\surd}$	&${\surd}$ \\
Graduates Degree			&Law Degree			&${\surd}$		&${\surd}$	&${\surd}$	&${\surd}$ \\		
Graduates Degree			&Juris Doctor		&${\surd}$		&${\surd}$	&${\surd}$	&${\surd}$ \\	
Mail Address State		&California			&${\surd}$		&			&${\surd}$	&${\surd}$ \\
Mail Address City			&Stanford			&				&			&			&${\surd}$ \\
Parent Institution		&Stanford University	&				&			&${\surd}$	&${\surd}$ \\
Tuition Measurement		&US Dollar			&${\surd}$		&${\surd}$	&			&${\surd}$ \\
Webpage Category			&WebPage			&${\surd}$		&${\surd}$	&			&${\surd}$ \\
\hline
\end{tabular}
\caption{Predicted facts by SVD, FFD+ and FFD (Analogy). ``${\surd}$'' stands for whether the prediction is correct.}
\label{tab:casestudy}
\end{table}

\begin{enumerate}
\item FFD (Analogy) can predict facts such as (``Located In'', ``Stanford'') and (``Mail Address City'', ``Stanford'') while other methods fail. It implies that this model can predict some relation with multiple possible tails.
\item Analogy+ outperforms SVD in general while fails to exceed FFD (Analogy). The reason is that it fails to predict some general facts like (``Located In'', ``USA'') or (``Tuition Measurement'', ``US Dollar''). This may due to the high scores given to some wrong facts.
\item The SVD model can only predict those facts whose relation and tail belong to the selected relation-tail pairs while Analogy+ and FFD (Analogy) can predict more facts.
\item SVD model prefers to predict some basic facts such as ``Located In'' and ``Tuition Measurement''. This is because those relations appear a lot of times in the training set and have limited possible tail entities. Therefore, it is easy for SVD model to make such prediction.
\end{enumerate}

\section{Conclusions and Future Work}

In this paper, we introduce a new task of fact discovery from knowledge base, which is quite important for enriching KBs. It is challenging due to the limited information available about the given entities for prediction. We propose an effective framework for this task. Experimental results on real-world datasets show that our model can effectively predict new relational facts. We also demonstrate that the feedback learning approach is useful for alleviating the issue of data sparsity for the head entities with few facts.

Facts discovery from knowledge base is essential for enriching KBs in the real world. Despite the fact that our work shows some promising results, there still remains some challenges:
(1) There exists much more internal information such as relational paths and external information such as text, figures and videos on the web, which can be used to further improve the performance.
(2) The feedback learning approach in this paper is to simply utilize those confident predicted relational facts to enhance the model. Reinforcement learning may help us dynamically select those informative and confident relational facts.

 \section*{Acknowledgments}
The work described in this paper is partially supported by grants from the Research Grant Council of the Hong Kong Special Administrative Region, China (Project Codes: 14203414) and the Direct Grant of the Faculty of Engineering, CUHK (Project Code: 4055093). Liu and Lin are supported by the National Key Research and
Development Program of China (No. 2018YFB1004503) and the National
Natural Science Foundation of China (NSFC No. 61572273, 61661146007).

\bibliography{naaclhlt2019}
\bibliographystyle{acl_natbib}
\end{document}